\documentclass[10pt,a4paper]{article}

\usepackage[utf8]{inputenc}
\usepackage[T1]{fontenc}
\usepackage[english]{babel}
\usepackage[a4paper,top=1.6cm,bottom=1.8cm,left=1.5cm,right=1.5cm,columnsep=0.7cm]{geometry}
\usepackage{setspace}
\usepackage{graphicx}
\usepackage{amsmath,amssymb}
\usepackage{booktabs}
\usepackage{caption}
\usepackage{subcaption}
\usepackage{enumitem}
\usepackage{xcolor}
\usepackage{hyperref}
\usepackage{lmodern}
\usepackage{url}
\usepackage{csquotes}
\usepackage{cite}

\hypersetup{
    colorlinks=true,
    linkcolor=blue,
    citecolor=blue,
    urlcolor=blue,
    pdftitle={Empathetic Motion Generation for Humanoid Educational Robots via Reasoning-Guided Vision--Language Diffusion Architecture},
    pdfauthor={Dr Sun}
}

\begin{document}
\onehalfspacing

\title{Empathetic Motion Generation for Humanoid Educational Robots\\
via Reasoning-Guided Vision--Language--Motion Diffusion Architecture}

\author{Fuze Sun, Jiaqi Huo, Lejing Dai, Xinyu Fan\\
\small Department of Computer Science and Engineering\\
\small University of Liverpool}
\date{} 

\maketitle

\begin{abstract}
This article suggests a reasoning-guided vision-language-motion diffusion framework (RG-VLMD) for generating instruction-aware co-speech gestures for humanoid robots in educational scenarios. The system integrates multi-modal affective estimation, pedagogical reasoning, and teaching-act-conditioned motion synthesis to enable adaptive and semantically consistent robot behavior. A gated mixture-of-experts model predicts Valence/Arousal from input text, visual, and acoustic features, which then mapped to discrete teaching-act categories through an affect-driven policy.These signals condition a diffusion-based motion generator using clip-level intent and frame-level instructional schedules via additive latent restriction with auxiliary action-group supervision. Compared to a baseline diffusion model, our proposed method produces more structured and distinctive motion patterns, as verified by motion statics and pairwise distance analysis. Generated motion sequences remain physically plausible and can be retargeted to a NAO robot for real-time execution. The results reveal that reasoning-guided instructional conditioning improves gesture controllability and pedagogical expressiveness in educational human-robot interaction.
\end{abstract}


\twocolumn

\section{Introduction}

\subsection{Background}
Educational robots have increasingly gained attention as embodied agents capable of supporting collaborative learning and personalized instruction \cite{chu2022artificial}, with prior work demonstrating its role in synchronizing with speech and supporting expressivity in humanoid robots \cite{wachsmuth2001lifelike}. Meanwhile, contemporary affective psychology provides computationally grounded models, such as multi-dimensional engagement vector \cite{article} and the OCC emotion model for robot \cite{kwon2007emotion} to infer user's mental state and engagement level with reference to multi-model interaction studies that indicates gestures and speech are simultaneously generated from a common cognitive source \cite{mcneill2008gesture} . However, most current systems still rely on pre-scripted reactive gesture libraries or heuristic controllers that fail to judge and reflect emotional or social nuance synchronously in motion \cite{dautenhahn2007socially}. This rigidity often results in unnatural interactions that appear mechanical, insensitive to learners' mental states, then disconnected them from the verbal or instructional context. It limits robots' effectiveness in fostering  meaningful interactions with emotion and empathy implemented as critical roles in instructional responses and learning motivations \cite{meyer2002discovering}. In addition, robots that cannot demonstrate genuine behaviors risks being perceived as rigid and impersonal, limiting the further implementation in educational domain.

With advancements in AI technologies, robots' capability to interact with humans in a more personalized and intuitive manner has evolved at a rapid pace \cite{obaigbena2024ai}. For instance, models that leverage automated semantic analysis from unannotated text input have shown promise in enabling corresponding gestures selection without manual linguistic annotation \cite{hartmann2005implementing}, together with probabilistic techniques further preventing unnatural determinism by involving variation across instances of repeated inputs\cite{stone2004speaking}. Specific progress in Vision-Language Models (VLMs) and Vision-Language-Action (VLA) architectures has introduced new possibilities for bridging perception, reasoning, and control in robotics. Frameworks such as OpenVLA \cite{kim2024openvla} and SmolVLA \cite{shukor2025smolvla} demonstrate that multi-modal transformers can convert textual instructions and visual context into structured action representations in both broad and specific world settings. These models facilitate robots to interpret tasks in a grounded, semantic manner, generating context-relevant reasoning-based motion primitives. However, most existing VLM and VLA systems focus on goal-oriented grasping or manipulation rather than affective or empathetic behavior, and themselves often lack the capability for real-time empathy reasoning required in educational human-robot interaction (HRI).

Motion generation driven by external data streams, such as synchronized speech \cite{wachsmuth2001lifelike} , has proven effective in facilitating context-aware expressivity. Parallel to this, Diffusion Policy has emerged as powerful motion generator for continues and high-dimensional control tasks. Models such as CDP \cite{ma2025cdp}, FDP \cite{patil2025factorizing} have proven that extended denoising diffusion process performs better than traditional policy learning or GAN-based approaches in stability, and sample diversity. These diffusion-based generators are particularly well fitted into humanoid motion synthesis tasks since they capture the natural variability of human movement while maintaining physical consistency. Nevertheless, current diffusion policies are largely blackbox, they produce smooth trajectories without any clear inference trace \cite{yang2023diffusion}, making it difficult for aligning their actions with social intent or pedagogical goals.

For a seamless Human Robot Interaction (HRI), Co-speech gesture generation focuses on synthesizing upper-body movements, such as those of the torso, hands, and face, in synchrony with speech to promote expressive and natural communication. Deep learning techniques such as Generative Adversarial Networks (GANs) \cite{goodfellow2020generative}, Varational Autoencoders (VAEs) \cite{kingma2013auto}, Vector Quantized VAEs (VQ-VAE), and diffusion-based  models are widely implemented to capture the underlying connections between speech signals and human motions. Recent studies have further introduced enhanced architectures, including VQ-VAE variants, cascaded processing frameworks, Transformer-based designs such as DiT, and diffusion-driven gesture synthesis approaches. In addition, more effective gesture dataset is collected by improved pose estimation methods \cite{tian2023recovering, zhang2021pymaf}, supporting data-driven learning for co-speech motion generation. These advancements provide a strong foundation for development a full-chain, co-speech motion generation framework capable of supporting context-adaptive and empathetic behaviour in educational scenarios with reference to psychological theory.

\subsection{Problem Statement \& Research Aim}
Current educational robots with motion capability often depends on pre-scripted gestures and open-loop controllers that unable to interpret learners' affective cues or synchronize motion with speech empathetically. This mismatch produces rigid, emotionally detached behaviors that weaken empathy, engagement, and instructional impact similar to issues previously identified in gestures systems relying on manually designed or deterministic behavior rules \cite{kipp2005gesture, wagner2014gesture} . Although recent advancements in VLMs and VLA systems integrate perception and reasoning for task execution, they remain limited in affective understanding and multi-modal alignment. Meanwhile, Diffusion based models have shown superior stability and diversity over traditional motion generation but lack of contextual and emotional grounding.

Furthermore, studies on speech-gesture coordination pointed out main categories of human gestures, such as, iconic, deictic, metaphorical, emblematic, and beat gestures, that contribute differently to communication and emphasize the importance of temporal synchrony between verbal and non-verbal cues for perceived consistency and naturalness during social interactions \cite{ng2010synchronized, chhatre2024emotional}. However, most existing robotic systems still rely on open-loop or manually aligned synchronization methods, often leading to disconnected speech and motion sequences that appear unnatural and insensitive to user intent \cite{habibie2021learning}.

To address these limitations, this research proposes a Reasoning-Guided Vision-Language-Motion Diffusion (RG-VLMD) framework specific to educational scenario. The framework integrates multi-model empathetic perception, interpretable text-based reasoning inspired by semantic-driven gesture selection approaches \cite{ao2023gesturediffuclip}, and FiLM-modulated diffusion policy influenced by probabilistic and iterative gesture refinement strategies that enhance natural variability while maintaining physical consistency \cite{yi2023generating, dosovitskiy2020image}. This model allows the robot to infer engagement and affect from educational context, reason about empathy intent and safety, then co-generate speech-aligned gestures that enhance social presence and learning outcomes. The goal is to realize an interpretable, reasoning-driven motion architecture that achieves empathy through synchronized multi-modal expression, facilitating future personalized, intelligent robot tutors.

\section{Related Work}

\subsection{Empathy in Human--Robot Interaction}
Empathy in Human-Robot Interaction (HRI) involves both cognitive empathy, encompassing the inference of users' beliefs and emotions, and affective empathy, categorized by reacting or sharing emotional states. In educational scenarios, empathetic behaviors contribute to improved rapport, learner motivations, and perceived support, since emotional understanding shapes how individuals interpret and respond to learning context \cite{schutz2006reflections}. Robotic systems integrated with emotional framework such as the OCC (Ortony, Clore, and Collins) model, have demonstrated the ability to recognize, interpret, and adapt to human emotions, prompting engagement during interaction, satisfaction, and trust \cite{kwon2007emotion}. Meanwhile, facial expressions remain crucial indicators for emotion recognition and appropriate empathetic response generation \cite{rawal2022facial}, and strategies such as Reactive Empathy and Parallel Empathy enable robots to align their affective responses with users' emotional states for better perceptions\cite{park2022empathy}. Even tiny empathetic cues, such as nodes,encouraging phrases, or basic facial expressions, have been shown to bring positive effects on student experience and motivation \cite{brown2014positive}, addressing the value of emotionally intelligent robot tutors in supporting  student development and learning \cite{alnajjar2021robots}.

However, most current empathetic systems in either virtual avatars or embodied agents apply predefined state-based strategies (e.g., "if confusion high, play encouraging animation") which constrain adaptability and long-horizon responsiveness. These methodologies incur disadvantages including limited motion expressivity, where empathy is rigidly expressed through isolated cues rather than coherent, full-body behaviors; discrete, non-adaptive action patterns that lack contextual and temporal modulation; and weak causality between empathy detection and motion generation, where emotion estimates act as triggers rather than shaping continues motion sequences. Empathetic cues are reconceptualized by the proposed framework as a continues control signal, generated from multi-modal perception input and complemented with an engagement state vector. This vector informs both reasoning process and continues body motion synthesis, facilitating context-sensitive, diverse responses that dynamically adaptive with the learner's need throughout the interaction.

\subsection{Social and Learning Theories}
Social Presence Theory (SPT) addresses that rich, contingent nonverbal behaviors, such as gestures and posture, enhance the user's subjective sense of "being with" a social other \cite{short1976social}. For HRI in educational settings, such cues are essential for mimicking attentiveness and fostering supportive learning environments. For example, consistent or responsive eye contact has been proven to strengthen the perceptions of attentiveness and engagement \cite{admoni2017social}, while nonverbal signals such as nods, gaze shifts, or body orientation help maintain rapport and interest by conveying empathy and understanding \cite{rodriguez2015bellboy}. These social cues also contribute to bidirectional communication, providing users a chance to infer the robot's intentions and inner states for better SPT satisfaction \cite{fiore2013toward}. Emotional Contagion Theory (ECT) further suggests that congruent affective expressions, no matter gestural, facial, or postural, can positively shape the learner's affect and engagement level \cite{hatfield2011emotional}, together with evidence that empathetic nonverbal response can enhance interaction quality and student motivation \cite{sidner2005explorations}.

Based on the learning sciences, Self-Determination Theory (SDT) emphasizes autonomy support, relatedness, and competence as core drivers of intrinsic motivation \cite{ryan2024self}. What is more, the importance of minimizing extraneous load and guiding learner's attention through grounded multi-modal cues such as pointing gestures, beat gestures, and gaze direction, is highlighted by Cognitive Load Theory (CLT) for better study outcomes \cite{sweller1988cognitive}. Yet, besides frequent invocation of these theories, current robot tutors often implement them only at the level of verbal strategies, such as offering praise or fixed behavioral scripts, rather than grounding them in the continuous dynamics of whole-body motion. Therefore, natural synchronization between speech and gesture remains coarse, limiting the robot's ability to direct attention effectively, sustain social presence, or reinforce relatedness though non-verbal behavior.

Our framework directly implements these theories at the motion generation level. Multi-modal empathetic reasoning determines when and how speech and gestures should be expressed, while a diffusion-based motion policy realizes these decisions as smooth, time-synchronized co-speech body trajectories. This structure enables the robot tutor to support attention management consistent with CLT, boost relatedness and autonomy support aligned with SDT, and improve social presence through cohesive, context-sensitive nonverbal communication.

\subsection{Reasoning in Vision-Language Models and Vision-Language-Action Systems}
Recent advancements in Vision-Language Models (VLMs) have dramatically expanded robots' capacity to interpret and reason over multi-modal input. Vision Transformer (ViT) as the foundational development \cite{dosovitskiy2020image}, introduced patch-based tokenization for visual understanding that enables large-scale pretraining and strong generalization. Building upon this, cross-modal alignment models, most prominently CLIP \cite{radford2021learning}, represents images and text in a shared latent space via contrastive learning over hundreds of millions of image-text pairs, facilitating a powerful paradigm for zero-shot multi-modal understanding. These ideas pave the way for a family of improved VLMs, such as OWL-ViT \cite{minderer2022simple} for open-vocabulary detection, FLIP \cite{yao2022guansong} for token-level cross-modal alignment, MASKCLIP \cite{zhou2022extract} for locally consistent masked modeling, together with the BLIP series \cite{li2022blip} which suggested efficient multi-modal pretraining utilizing Q-Former and Perceiver Resampler modules. Meanwhile, vision foundation models such as SAM / SAM2 \cite{kirillov2023segment, ravi2024sam} extended zero-shot generalization to segmentation and video-level consistency, showcasing the increasing capability of multi-modal systems to execute structured reasoning over rich perceptual channels and scenes, which act as an essential prerequisite for socially intelligent robots.

Building upon these perception and reasoning advances, Vision-Language-Action (VLA) systems aim to align high-level reasoning with robot hardware control. For instance, RT-1 \cite{brohan2022rt}, OpenVLA \cite{kim2024openvla}, and Octo \cite{team2024octo} as end-to-end models, map visual-textual inputs directly to actions. This typically produces low-dimensional motor commands but lacks interpretable intermediate representations such as gesture intent, empathy cues, or proxemics rules. Emerging hierarchical approaches, such as VoxPoser \cite{huang2023voxposer}, ReKep \cite{huang2024rekep}, and Code-as-Policies (COPA) \cite{liu2024rl}, introduce semantic layers such as affordance maps, relational keypoints, or executable task code, yet hold largely task-specific, focusing on manipulation and navigation rather than socially grounded body behavior. Consequently, only limited support for pedagogical reasoning, co-speech gesture synchrony, or affect-aware motion, is provided by current VLM/VLA, but all of which are essential for educational HRI. These disadvantages motivate our proposed Reasoning-Guided VLM Architecture, which compress empathetic intent, gesture semantics, and contextual constraints as explicit structured plans to drive downstream motion generation, facilitating context-adaptive, expressive empathetic behavior not achieved by existing paradigms.

\subsection{Diffusion-based Motion Generation}
Diffusion models have emerged as strong generative models for human motion, outperforming GANs and autoregressive transformers in realism, diversity, AND temporal coherence. Current speech-driven gestures systems such as DiffuseStyleGesture \cite{yang2023diffusestylegesture}, DiffGesture \cite{zhu2023taming}, DiffSHEG \cite{chen2024diffsheg}, and AMUSE \cite{chhatre2024emotional} demonstrate that diffusion policy can generate speech-synchronized, smooth human-like gestures with controllable style and emotional tone. They provide valuable building blocks, such as DiffuseStyleGesture enables stylisitc modulation of co-speech motion; DiffGesture proposed a strong baseline for audio-conditioned natural gestures; DiffSHEG supports real-time speech-oriented holistic expression; and AMUSE disentangles affective content for emotionally congruent motion. Parallel researches in Diffusion Model for robot control \cite{ma2025cdp, patil2025factorizing} showcases that extended denoising process can handle continues, diverse, and high-dimensional action spaces. Together, Diffusion's strength is highlighted by these advances as a generative prior for smooth, human-like motion.

Nevertheless, existing diffusion-based gesture generation systems remain limited for empathetic HRI in educational scenarios. They mainly operate in manipulation space, without modeling diverse joint limits, torque bounds, or proxemics, reducing compatibility for safety-critical classroom settings. Input signals for conditioning purpose are typically restricted to audio, style token, or manual emotion labels, providing little psychological interpretability and no mechanism to align motion with empathetic intent, engagement shifts, or pedagogical dynamics. These systems are also weakly paired to real-time multi-modal perception and can not adapt to learner's changing mental states. Even worse, they implement a single generative stream that lack a reactive override for rapid flinch/avoidance responses through an executive to arbitrate between planned and reflective motions. The RG-VLMD framework proposed in this paper addresses these disadvantages through reasoning-guided FiLM conditioning, a deliberative-executive-reactive architecture, and a parallel reactive emotion pathway, enabling safe and psychologically grounded human-like motion for educational robots.

\section{Proposed Architecture and Methodology}
\subsection{Overall Architecture}
The proposed Reasoning-Guided Vision-Language-Motion Diffusion (RG-VLMD) framework is orchestrated as a three-layer pipeline that transformers multi-modal perception into pedagogically grounded speech content and synchronized tutoring motion. This architecture integrates continues affective estimation, discrete instruction planning, and diffusion-based motion generation into a unified control loop.

Perception layer: The MOSEI-trained predictor (Section 3) is used by the system to first process audio, textual, and visual interaction signals to estimate the learner's affective state in continues Valence-Arousal (V/A) space. Smoothed affective values and short-term trends provide a stable representation of the learner's emotional and engagement dynamics.

Pedagogical reasoning layer: The affect-driven policy maps the continues affect state to a discrete teaching-act label (e.g. praise, hint, explain). This teaching-act serves as the interpretable intent representation and is shared throughout modalities: it conditions on the Large Language Model (LLM) to generate pedagogical aligned speech and synchronously guides gesture synthesis.

Motion generation layer: Given input speech features and the selected teaching-act label, the RAPID-Motion diffusion model produces co-speech gesture trajectories in bvh file. The simultaneous motion is retargeted to the NAO robot under physical constraints, ensuring stable, temporally aligned movements.

\begin{figure*}[t]
\centering
\includegraphics[width=\textwidth]{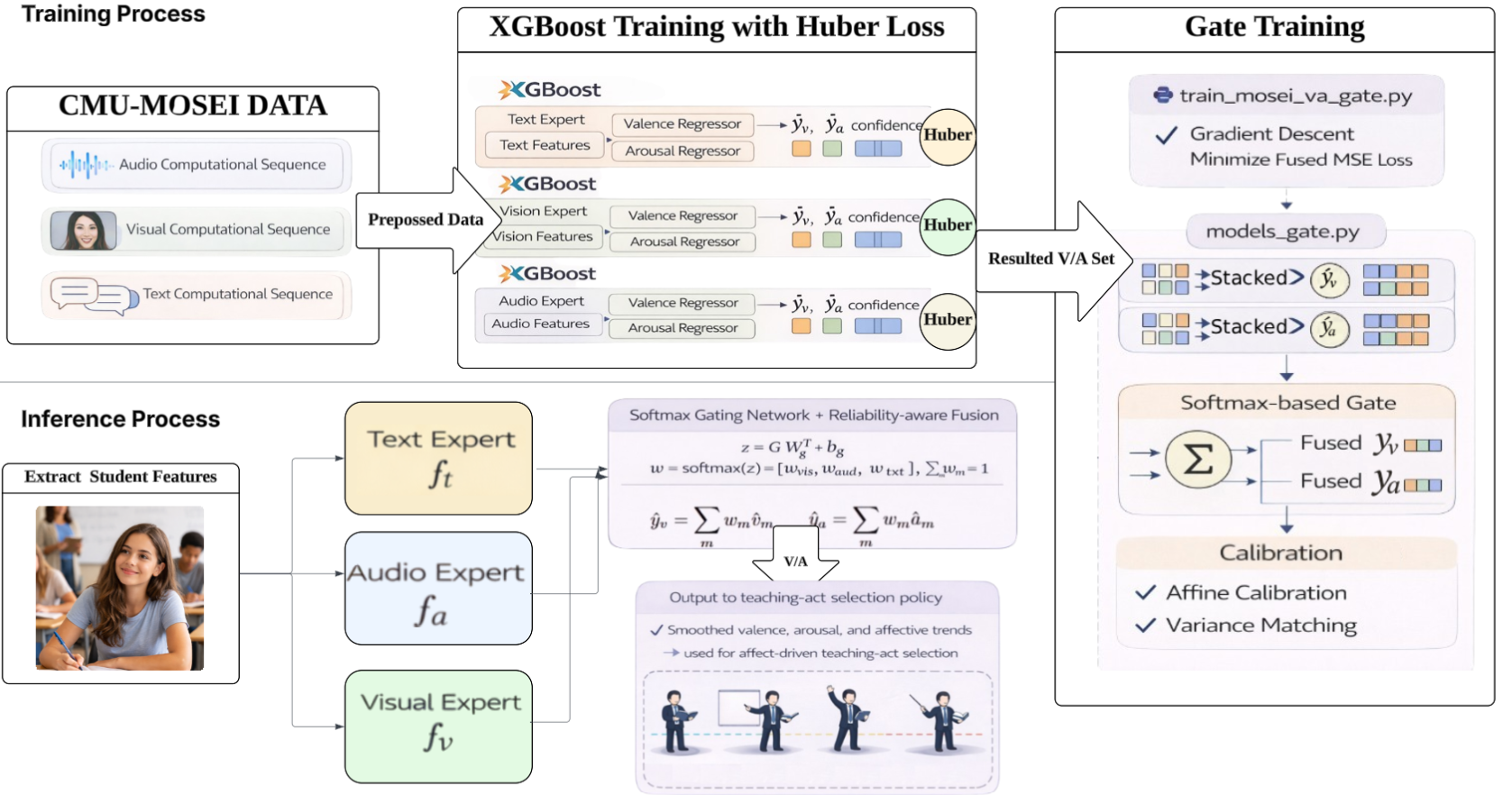}
\caption{
Training and inference flow of the proposed valence/arousal estimator. 
During training, modality-specific XGBoost experts are first optimized on CMU-MOSEI using Huber-style regression objectives, and their outputs are then used to train a softmax-based reliability gate with fused mean squared error loss. 
During inference, student text, visual, and acoustic features are processed by the trained modality experts, followed by reliability-aware gated fusion and calibration to produce final valence and arousal estimates for downstream teaching-act selection.
}
\label{fig:va_estimator_flow}
\end{figure*}

\subsection{Affective Perception via Valence–Arousal Estimation on CMU-MOSEI}
A continues and computational grounded representation of the learner's affective state is required by empathetic speech \& motion generation. Instead of counting on discrete emotion categories, which introduce artificial boundaries and limited control resolution, this work adopts the Valence/Arousal (V/A) dimensional model of affect. From this representation, valence captures affective polarity (negative to positive) and arousal reflects activation intensity (calm to excited). The dimensional formulation aligns well with affective psychology and is particularly helpful for continues control in generative speech \& motion systems, where gradual modulation of expressivity and energy is required.

For training a robust affective estimator, we employ the CMU-MOSEI corpus, a large scale and well-structured multi-modal dataset containing aligned textual, visual, and acoustic segments annotated for sentiment and emotion. Supervised regression of continuous affective dimensions from heterogeneous perceptional signals is enabled by its multi-view structure.

\subsubsection{Problem Formulation}
Three modality-specific feature representations are extracted for each aligned segment i:
\begin{equation}
X_t^{(i)} \in \mathbb{R}^{600}, \quad
X_v^{(i)} \in \mathbb{R}^{1426}, \quad
X_a^{(i)} \in \mathbb{R}^{148}
\end{equation}
corresponding to textual embedding, facial/visual descriptors, and acoustic features, respectively.

The targets for regression are continues affect annotation:
\begin{equation}
y_v^{(i)} \in [-1,1], \qquad
y_a^{(i)} \in [0,1]
\end{equation}
representing refactored valence and arousal.

What is more, a reliability descriptor
\begin{equation}
G^{(i)} \in \mathbb{R}^{3}
\end{equation}
is computed for each segment to label normalized modality confidence (e.g. signal quality or detection reliability). It does not directly influence the modality regressors, but serves as an input to the fusion gate, enabling dynamic arbitration among modalities.

The learning objective is to approximate:
\begin{equation}
f : (X_t, X_v, X_a, G) \rightarrow (\hat{y}_v, \hat{y}_a)
\end{equation}

where $(\hat{y}_v, \hat{y}_a)$ denote predicted valence and arousal.

\subsubsection{Two-Stage Reliability-Aware Mixture-of-Experts}
We adopt a two-stage gated mixture-of-experts architecture (see Figure ~\ref{fig:va_estimator_flow}) to achieve high predictive accuracy while maintaining computational efficiency suitable for real-time human robot interaction.

Stage A: Modality-Specific Nonlinear Experts
Three independent gradient-boosted regression models are trained as one per modality:

\begin{align}
f_t &: X_t \rightarrow (\hat{v}_{txt}, \hat{a}_{txt}) \\
f_v &: X_v \rightarrow (\hat{v}_{vis}, \hat{a}_{vis}) \\
f_a &: X_a \rightarrow (\hat{v}_{aud}, \hat{a}_{aud})
\end{align}

Every expert algorithm is implemented using XGBoost with a pseudo-Huber regression objective, providing robustness to label outliers and noise. A large estimator budget together with early stopping enforce expressive nonlinear modeling without overfitting. Different random seeds are used across modalities to reduce correlated bias among experts.

Importantly, predictions of modality remain unclipped during training and fusion, preserving full dynamic range. Empirically, compressing variance issue was found after early clipping, particularly for arousal, therefore weakening the gate's ability to learn distinctive weighting patterns.

Stage B: Softmax Gating Network
A lightweight linear gating network dynamically determines modalities' weights references to reliability features G. The gate computes:

\begin{equation}
z = G W_g^\top + b_g
\end{equation}

\begin{equation}
w = \text{softmax}(z)
\end{equation}

where

\begin{equation}
W_g \in \mathbb{R}^{3 \times 3}, \quad
b_g \in \mathbb{R}^{3}
\end{equation}

and

\begin{equation}
w = [w_{vis}, w_{aud}, w_{txt}], \qquad
\sum_{m} w_m = 1.
\end{equation}
The final fused predictions are calculated based on convex combination:

\begin{equation}
\hat{y}_v = \sum_{m} w_m \hat{v}_m
\end{equation}

\begin{equation}
\hat{y}_a = \sum_{m} w_m \hat{a}_m
\end{equation}

Reliability-aware modality arbitration is enabled through this formulation. For example, if visual tracking confidence decreases, the corresponding reliability input reduces $w_{vis}$, allowing acoustic or textual channels to dominate.

The parameters from gate are optimized through gradient-based minimization of regression loss on fused predictions, optionally restricted by an entropy term to prevent premature modality collapse. Since its shallow structure (linear+softmax), the gate keeps light-weight and interpretable, facilitating real-time inference in real-world human-robot interaction scenario.

\subsection{Affect-Driven Teaching-Act Selection Policy}
We introduce an affect-driven teaching-action selection module to bridge the gap between continues affect perception and discrete pedagogical control. This component pairs sentence-level Valence/Arousal estimates during a conversation into one of eight pedagogical categories:
\[
\{\textit{praise}, \textit{hint}, \textit{explain}, \textit{checkin}, \textit{slow\_down}, \textit{challenge}, \textit{neutral}, \textit{unclear}\}.
\]

Different from the regex-based labeling mechanism used during diffusion dataset preprocessing to generate pseudo action annotations, this adaptive policy solely relies on the predicted affective dynamics. The goal is to produce a stable, psychologically grounded control signal that regulates both verbal and non-verbal instructional behavior in the educational setting.

Affective State Representation
For each sentence i from the conversation, the Valence-Arousal Estimation Model predicts valence $v_i \in [-1,1]$ and arousal $a_i \in [0,1]$. To reduce jitter and highlight temporal evolution, we apply exponential smoothing:

\begin{equation}
\bar v_i = \alpha v_i + (1-\alpha)\bar v_{i-1}, \qquad
\bar a_i = \alpha a_i + (1-\alpha)\bar a_{i-1},
\end{equation}

where $\alpha = 0.3$. Short-term affective trends are computed as:
\begin{equation}
\Delta v_i = \bar v_i - \bar v_{i-1}, \qquad
\Delta a_i = \bar a_i - \bar a_{i-1}.
\end{equation}

Since predicted V/A values show compressed dynamic ranges relative to theoretical bounds, decision thresholds are calibrated using empirical percentiles of our development set outcomes rather than fixed theoretical cutoffs. In this case, valence values above approximately 0.24 correspond to obvious positive effect, values below -0.11 indicate negative affect, and arousal values above approximately 0.42 correspond to high activation within the model's output distribution.

Pedagogical Decision Logic
This mapping follows a prioritized, safety-aware policy:
\begin{enumerate}[leftmargin=1.2em]
\item \textbf{Regulation First.} 
If arousal is high and valence is negative, or if arousal increases while valence drops down, the system picks slow\_down to prevent escalation.

\item \textbf{Positive Reinforcement and Progression.}  
Strong positive valence leads to praise. Positive valence combined with elevated arousal and improving trend invokes challenge, encouraging advancement.

\item \textbf{Instructional Support.}  
Hint is resulted by a negative valence trend without high arousal. Neutral valence together with moderate arousal defaults to explain, providing structured guidance from the robot tutor.

\item \textbf{Stable Neutral States.}  
Neutral valence and low arousal yields neutral.

\item \textbf{Uncertainty Handling.}  
The system emits unclear when affect prediction are unreliable (e.g. low confidence).

\end{enumerate}
This hierarchical formulation ensures that emotional restriction precedes pedagogical escalation, aligning educational intent with leaner well-being.

\subsubsection*{Dual Conditioning Role}
The selected teaching act serves as a shared control variable across the whole architecture. Specifically:
\begin{itemize}[leftmargin=1.2em]
\item The Large Language Model (LLM) is conditioned by it to generate pedagogical aligned instructional speech (e.g. explanatory, encouraging, challenging tone).
\item It conditions the RAPID-Motion diffusion model as a discrete control vector, regulating motion style through global act embedding and frame-level FiLM modulation.

Through mapping continues affective dynamic into structured pedagogical intent, this module establishes a coherent interface between perception, reasoning, speech generation, and motion synthesis. It thus bridges the gap between affect and action within the proposed Reasoning-Guided Vision-Language-Motion Diffusion Framework.

\end{itemize}

\begin{figure*}[t]
\centering
\includegraphics[width=\textwidth]{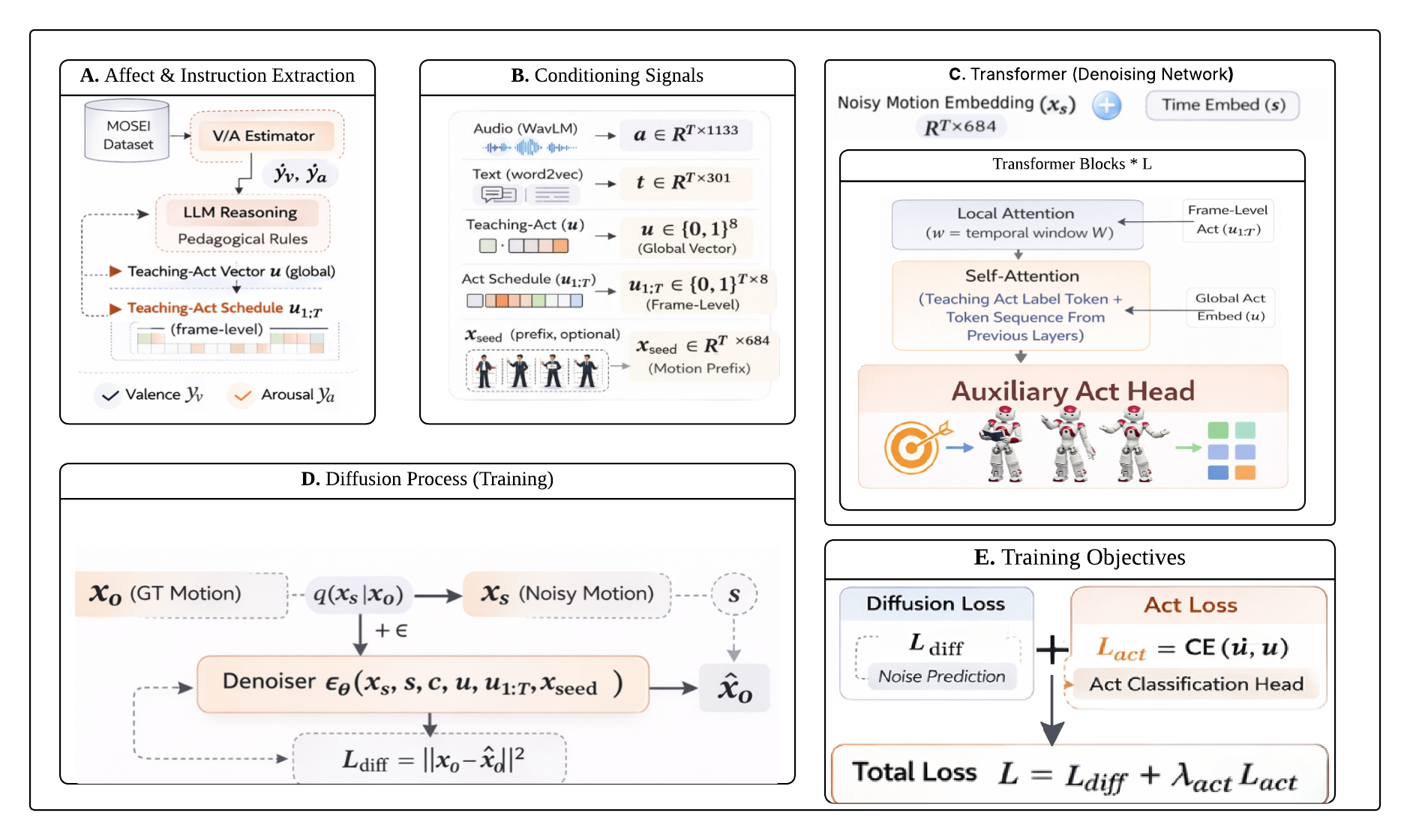}
\caption{
\textbf{RAPID-Motion diffusion framework for pedagogical co-speech gesture generation.}
Student affect is estimated using a MOSEI-trained Valence–Arousal model and converted by an LLM reasoning module into a clip-level teaching-act vector $u$ and frame-level schedule $u_{1:T}$. 
Audio and text embeddings are used as multimodal conditioning signals, together with instructional vectors and an optional motion prefix.
A transformer-based diffusion denoiser generates motion tokens using local attention and self-attention. Training minimizes diffusion reconstruction loss with an auxiliary act-classification loss to enforce pedagogical consistency.
}
\label{fig:rapid_motion_architecture}
\end{figure*}

\subsection{RAPID-Motion: Reasoning-Aware Pedagogical Instructional Diffusion for Motion}

\subsubsection{Overview and Objective}
We propose RAPID-Motion (Reasoning-Aware Pedagogical Instructional Diffusion for Motion) to generate empathetic, speech-synchronized humanoid motion grounded in pedagogical reasoning (see Figure ~\ref{fig:rapid_motion_architecture}) . RAPID-Motion is a teaching-act-conditioned diffusion policy that generates continues co-speech gesture trajectories while incorporating educational instructions explicitly.

Different from conventional co-speech diffusion systems that primarily condition on text or audio features, RAPID-Motion integrates pedagogical reasoning signals derived from the upstream affect estimation model. Specifically, continues Valence/Arousal predictions obtained from the affect model trained on MOSEI-Dataset inform a LLM reasoning process that produces structured teaching-act control vectors. These vectors work as high-level semantic instructions that guide teaching motion generation. In this way, synthesis of gesture becomes explicitly grounded in inferred leaner affective state and instructional strategy rather than being conditioned only by plain text or acoustic prosody.

Formally, RAPID-Motion models the conditional distribution:
\begin{equation}
p_{\theta}\big(x \mid c, u, u_{1:T}, x_{\mathrm{seed}}\big),
\end{equation}

where x represents the target denoised motion sequence, c denotes multi-modal speech conditioning (audio and text embedding), u is a clip-level teaching-act vector, $u_{1:T}$ is a time-varying teaching-act schedule, and xseed is an optional motion prefix used to ensure temporal continuity.

\subsubsection{Data Representation and Conditioning Signals}
For a motion clip of length T frames, the aligned multi-modal data streams from BEAT are defined as follows:

The motion sequence
\begin{equation}
x \in \mathbb{R}^{T \times D_m}, \quad D_m = 684
\end{equation}
represents the parameterization of the BRAT v0 motion.

The acoustic features are extracted using a WavLM-based representation and the textual features correspond to word2vec embeddings aligned to the same temporal frames.

\begin{equation}
a \in \mathbb{R}^{T \times D_a}, \quad D_a = 1133
\end{equation}

\begin{equation}
t \in \mathbb{R}^{T \times D_t}, \quad D_t = 301
\end{equation}

Pedagogical intent is encoded at 2 hierarchical levels:
\begin{equation}
u \in \{0,1\}^{K}, \quad K=8
\end{equation}

\begin{equation}
u_{1:T} \in \{0,1\}^{T \times K}.
\end{equation}

Vector u represents the dominant instructional strategy (e.g. explanation, challenge, hint), while $u_{1:T}$ enables dynamic pedagogical modulation across time. 

\subsubsection{Diffusion Architecture and Objective}
RAPID-Motion implements a denoising diffusion formulation over motion sequences. Given a clean motion clip $x_0$, a diffusion timestep s is sampled uniformly, and Gaussain noise is added to get a noisy motion representation xs.

The denoiser is trained to predict the injected noise by minimizing:
\begin{equation}
\mathcal{L}_{\mathrm{diff}} =
\mathbb{E}_{x_0,\epsilon,s}
\left[
\|\epsilon - \epsilon_{\theta}(x_s, s, c, u, u_{1:T}, x_{\mathrm{seed}})\|_2^2
\right].
\end{equation}

The network outputs motion only. Conditioning signals influence hidden representations via attention and modulation mechanisms but are not reconstructed.

The Diffusion Model implements transformer as the denoiser operating on frame-level tokens. All inputs are projected into a shared latent space of dimension d:

\begin{itemize}[leftmargin=1.2em]
\item Motion tokens: $\mathbb{R}^{D_m} \rightarrow \mathbb{R}^{d}$,
\item Conditioning tokens: $\mathbb{R}^{1434} \rightarrow \mathbb{R}^{d}$,
\item Global act embedding: $\mathbb{R}^{K} \rightarrow \mathbb{R}^{d}$,
\item Diffusion timestep embedding: $s \rightarrow \mathbb{R}^{d}$.
\end{itemize}

Cross-attention enables motion tokens to attend to speech-conditioning tokens, preserving temporal alignment between verbal content and gesture dynamics. To balance computational efficiency and temporal fidelity, RAPID-Motion employs a local-attention mechanism. Instead of allowing global attention across the whole sequence, attention is fixed to a constant temporal window around each frame. 

If W represents the window size, tokens at time t attend only to conditioning tokens range in [t-W, t+W]. This design reduces quadratic complexity while strengthening local audio-gesture synchrony. Sequence lengths are adjusted via cropping or padding to satisfy window constraints during training

\subsubsection{Pedagogical Conditioning and Supervision}
A distinctive characteristic of RAPID-Motion is explicit pedagogical conditioning. 

Global Instructional Control: The overall communication style of the motion sequence is governed by the clip-level vector u. It encodes high-level educational intent and influences gesture expansiveness, amplitude, and expressive tone.

The proposed model introduces frame-level instructional conditioning together with overall intent labeling to support dynamic pedagogical behavior within a single utterance. Unlike FiLM-based modulation, instructional signals are incorporated through latent additive conditioning within the diffusion-based motion generator to learn the connection between action sequence and intention. Let $h_t$ represent the latent motion embedding at frame t. The condition representation is computed as:

\begin{equation}
h'_t = h_t + \lambda_c e_c + \lambda_f e_t ,
\end{equation}

where $e_c$ is the embedding of clip-level instruction label, $e_t$ is the embedded frame-level act vector, and $\lambda_c$ and $\lambda_f$ are scaling factors controlling the contribution of global and local instructional context respectively.

RAPID-Motion incorporates an auxiliary act-classification head to ensure that internal representations encode pedagogical intent into different clusters rather than ignoring the act-conditioning pathway. A clip-level act label is predicted from intermediate transformer representations and optimized through cross-entropy loss:

\begin{equation}
\mathcal{L}_{\mathrm{act}} = \mathrm{CE}(\hat{u}, u).
\end{equation}

The total training objective becomes:

\begin{equation}
\mathcal{L} =
\mathcal{L}_{\mathrm{diff}}
+ \lambda_{\mathrm{act}} \mathcal{L}_{\mathrm{act}},
\end{equation}

where $\lambda_{\mathrm{act}}$ balances diffusion reconstruction and instructional supervision

This auxiliary head encourages the diffusion model to maintain action-distinctive latent features, preventing condition collapse.
\subsubsection{Training and inference Pipeline}
Train: Fixed-length motion windows of T frmaes are sampled from aligned clips during training. Multi-model conditioning is constructed by concatenating frame-aligned text and acoustic embeddings, forming $c = [a \parallel t]$. The corresponding clip-level teaching-act vector u and frame-level schedule $u_{1:T}$ are retrieved, and an motion prefix x-seed is provided when seed-based continuation is enabled.

The diffusion timestep s is sampled consistently, and Gaussian noise is injected to the clean motion sequence to generate a noisy sample  $x_s$. The transformer-based denoiser's input is in $(x_s, s, c, u, u_{1:T}, x_{\mathrm{seed}})$ format and predicts the added noise. The diffusion model is optimized via the reconstruction objective combined with auxiliary pedagogical supervision:

\begin{itemize}[leftmargin=1.2em]
\item The diffusion loss encourages accurate recovery of motion trajectories.
\item The auxiliary act-classification loss enforces pedagogical grounding in the latent representation.
\end{itemize}

Each sampled window applies cross-local attention to maintain speech-gesture alignment while ensuring computational efficiency. The loss is primarily implemented to the non-seed region to promote stable continuation behavior when seed prefixes are used.

\begin{table*}[t]
\centering
\caption{Normalized motion statistics computed from BVH trajectories under different teaching-act conditions. Values are normalized by the maximum across all conditions.}
\label{tab:motion_stats}
\begin{tabular}{lcccccl}
\hline
Teaching-Act & Amplitude & Velocity & Jerk & Energy & Range & Interpretation \\
\hline
explain   & 1.00 & 1.00 & 1.00 & 1.00 & 1.00 & highly expressive explanatory motion \\
checkin   & 0.50 & 0.50 & 0.14 & 0.07 & 0.80 & moderate monitoring gestures \\
challenge & 0.47 & 0.47 & 0.04 & 0.02 & 0.84 & controlled but directive motion \\
neutral   & 0.45 & 0.45 & 0.14 & 0.04 & 0.85 & regular conversational motion \\
praise    & 0.34 & 0.34 & 0.03 & 0.02 & 0.90 & smooth congratulatory gestures \\
unclear   & 0.61 & 0.61 & 0.04 & 0.03 & 0.53 & irregular / uncertain motion \\
\hline
\end{tabular}
\end{table*}

\subsection{Datasets and Data Augmentation}
CMU-MOSEI for Valence/Arousal Estimation: We train and evaluate the affect perception module on the CMU-MOSEI corpus which is a large-scale SOTA multi-modal benchmark containing aligned language, visual, and acoustic computational sequences with sentiment and emotion annotations (refactored to V/A). To ensure the preprocessing and reproducibility are consistent, we acquire and align MOSEI using  CMU Multi-Modal SDK (mmsdk), which provides standardized data streams and alignment utilizes across modalities.

For each aligned sentence segment, we extract modality-specific features (text, vision, audio) and train a lightweight two-stage mixture-of-experts predictors as described in Section 3.1.  The official MOSEI split protocol is followed to reserve a development set for calibration (affine calibration and variance matching) and a held-out test set for reporting final performance.

Since our affect estimation model is designed for robust real-world deployment, we apply simple label-preserving perturbations during training and validating-time stress tests: (i) randomly masking one modality's features to simulate camera/audio degradation, and (ii) EMA on predicted V/A to reduce jitter in downstream control. These augmentations are implemented on the feature and decision level rather than raw media to remain compatible with SDK-provided representations.

BEAT for Co-Speech Motion generation: To synthesis motions, we train RAPID-Motion on the BEAT dataset which provides high-quality motion capture gesture sequences paired with text and audio features. BEAT is suitable for data-driven co-speech gesture learning since it contains approximately 76 hours of motion capture recordings from 30 speakers across multiple scenarios.

An HDF5-based processed representation consistent with pipelines popularized in diffusion-based gesture system (audio + text-alignment, fixed-rate motion features, and clip-wise indexing) is used to support diffusion training at scale and simplify I/O. As a methodological base and engineering reference for diffusion training and evaluation practice, we adapt established gesture conversion pipelines suggested in DiffuseStyleGesture+ (e.g. multi-modal conditioning, seed prefixing, and efficient data packaging).

Teaching-act supervision for motion training. During training only, we extract and attach teaching-act labels to each clip to train RAPID-Motion's action-conditioned generation. At runtime, teaching-act labels are provided by the affect-to-act controller (Section 3.2) rather than by text-derived heuristics.


\section{Implementation and System Evaluation}

\subsection{Model Implementation}
The proposed Reasoning-Guided Vision-Language-Motion Diffusion (RG-VLMD) framework is implemented in Python and integrated with a humanoid robot platform (NAO) for real-time interaction in educational scenarios. The overall control stack is in a modular architecture consisting of three main components: multi-modal affect perception, pedagogical reasoning, and diffusion-based empathetic motion generation.

The affective perception module runs on an external computing node and estimates the learner's emotional state using a multi-modal Valence/Arousal predictor trained on the CMU-MOSEI dataset. The continues valence and arousal scores that represent the learner's inner state are produced based on the synchronized input text, audio, and visual features.

The pedagogical reasoning module operates on a GPU server and interprets the estimated affective states using a Large Language Model (LLM). The affective signals are converted into structured instructional strategies represented as teaching-act vectors by the LLM. These vectors provide high-level semantic guidance for downstream gesture generation.

\begin{figure}[t]
\centering

\newcommand{\imgH}{3.2cm}

\begin{subfigure}[t]{0.32\linewidth}
\centering
\includegraphics[width=\linewidth,height=\imgH,trim=20 20 20 20,clip]{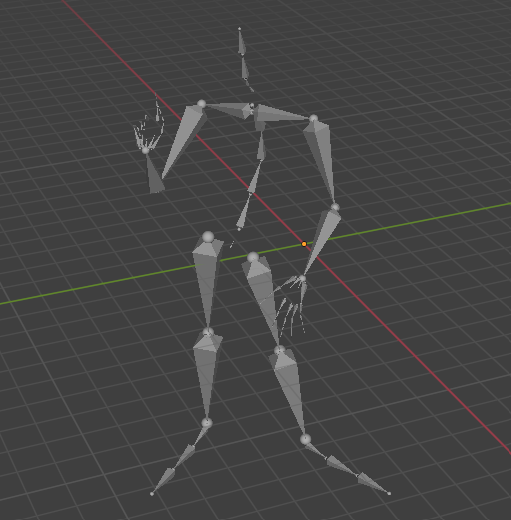}
\caption{Challenge}
\end{subfigure}
\begin{subfigure}[t]{0.32\linewidth}
\centering
\includegraphics[width=\linewidth,height=\imgH,trim=20 20 20 20,clip]{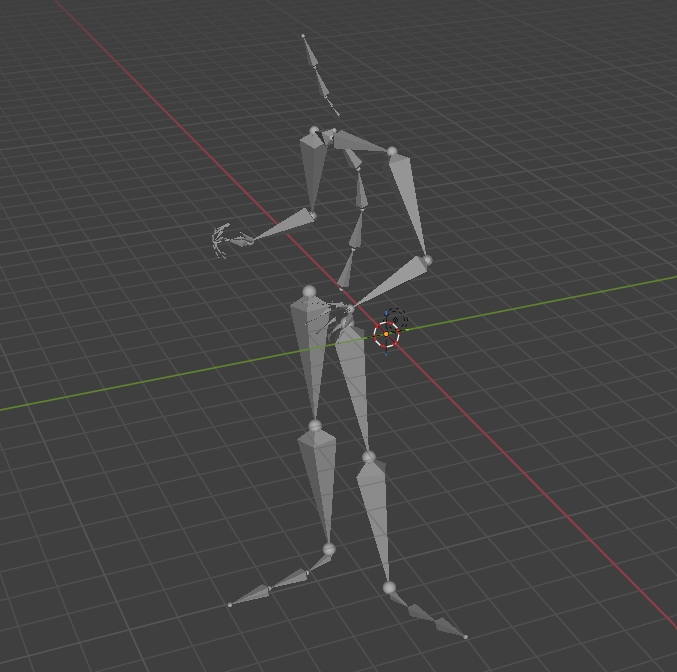}
\caption{Check-in}
\end{subfigure}
\begin{subfigure}[t]{0.32\linewidth}
\centering
\includegraphics[width=\linewidth,height=\imgH,trim=20 20 20 20,clip]{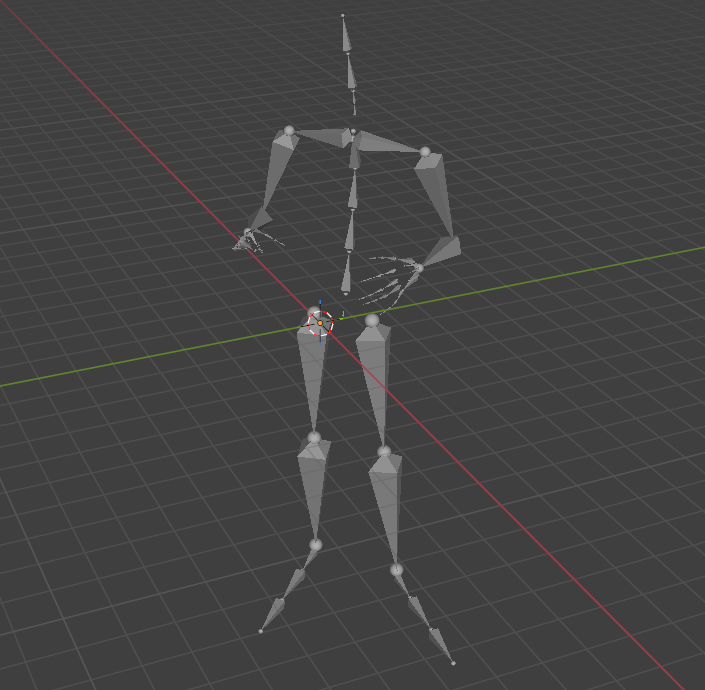}
\caption{Explain}
\end{subfigure}

\begin{subfigure}[t]{0.32\linewidth}
\centering
\includegraphics[width=\linewidth,height=\imgH,trim=20 20 20 20,clip]{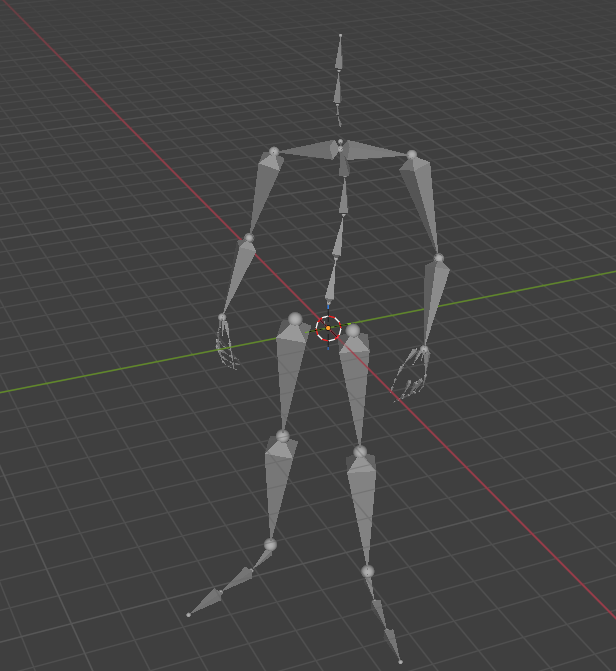}
\caption{Neutral}
\end{subfigure}
\begin{subfigure}[t]{0.32\linewidth}
\centering
\includegraphics[width=\linewidth,height=\imgH,trim=20 20 20 20,clip]{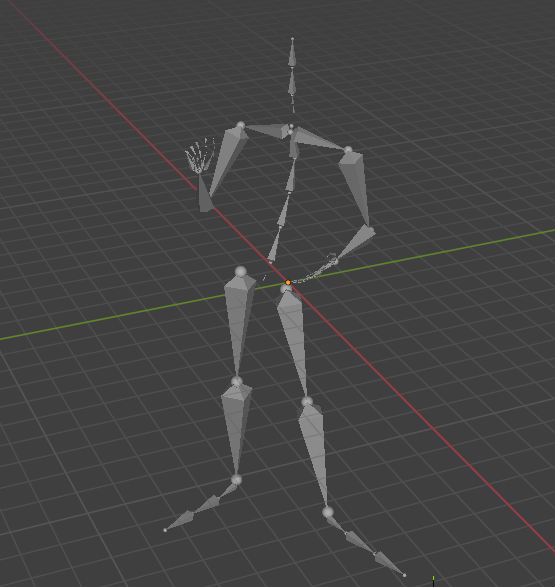}
\caption{Praise}
\end{subfigure}
\begin{subfigure}[t]{0.32\linewidth}
\centering
\includegraphics[width=\linewidth,height=\imgH,trim=20 20 20 20,clip]{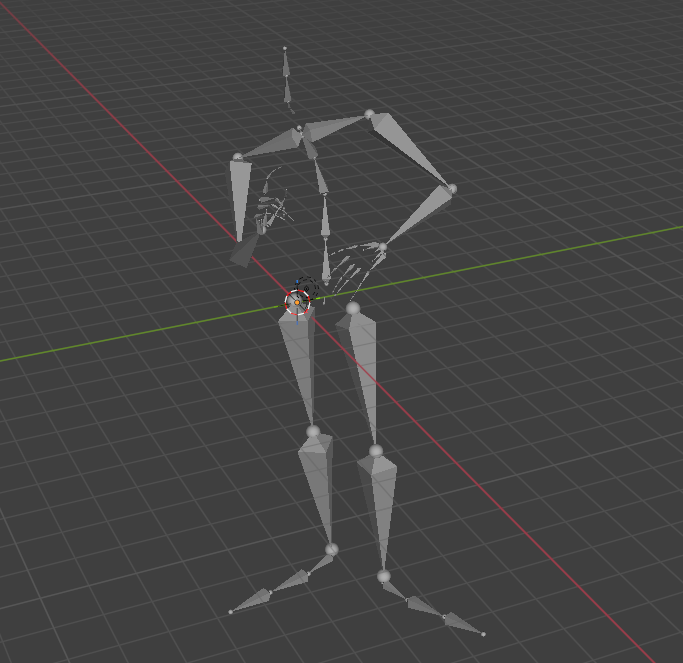}
\caption{Unclear}
\end{subfigure}

\caption{
Representative gesture poses generated by the proposed model under different teaching-act conditions in the simulation environment. 
The poses illustrate that the diffusion policy produces distinct motion styles corresponding to pedagogical intentions, including expressive gestures for explanation and praise, directive posture for challenge, and low-intensity motion for neutral interaction.
}

\label{fig:gesture_examples}

\end{figure}

The RAPID-Motion Diffusion Model produces co-speech gestures conditioned on multi-modal text/tone embeddings and pedagogical control signals. For realizing real-time interaction, the denoising process is executed with a reduced number of diffusion steps while maintaining gesture smoothness and expressiveness.

Finally, the generated motion trajectories in bvh files are subsequently retargeted to the NAO robot's joint space and executed through its controller, enabling simultaneous speech and gesture delivery during tutoring interactions.

\subsection{Module Validation}
Component-level validation is conducted to evaluate the reliability of each module in the proposed architecture.

\begin{figure}[t]
\centering

\begin{subfigure}[b]{0.48\columnwidth}
    \centering
    \includegraphics[width=\linewidth]{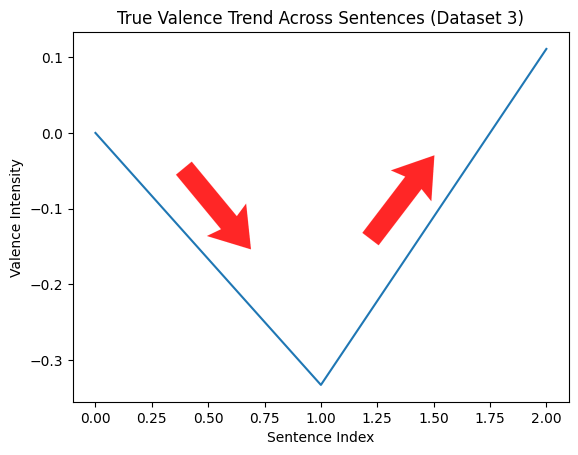}
    \caption{True valence (Dataset 3)}
\end{subfigure}
\hfill
\begin{subfigure}[b]{0.48\columnwidth}
    \centering
    \includegraphics[width=\linewidth]{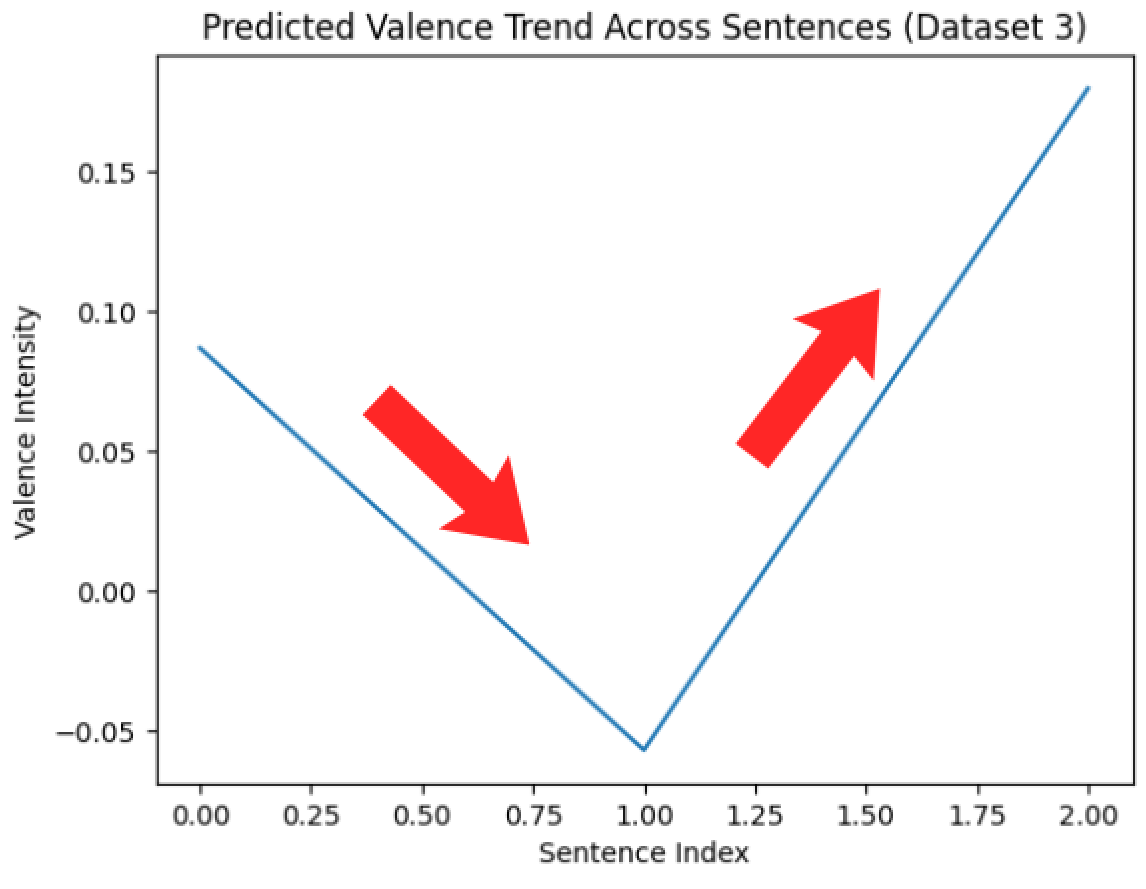}
    \caption{Predicted valence}
\end{subfigure}

\vspace{0.3em}

\begin{subfigure}[b]{0.48\columnwidth}
    \centering
    \includegraphics[width=\linewidth]{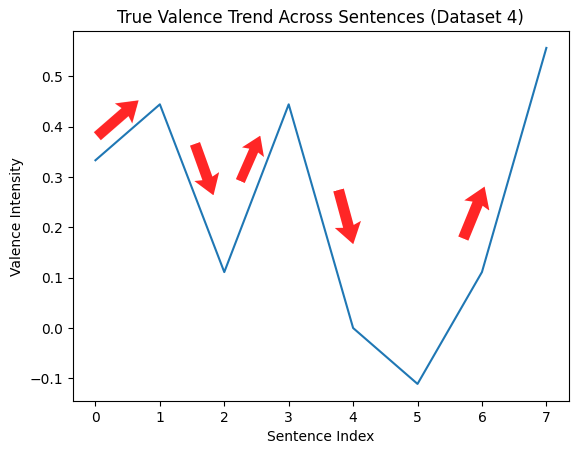}
    \caption{True valence (Dataset 4)}
\end{subfigure}
\hfill
\begin{subfigure}[b]{0.48\columnwidth}
    \centering
    \includegraphics[width=\linewidth]{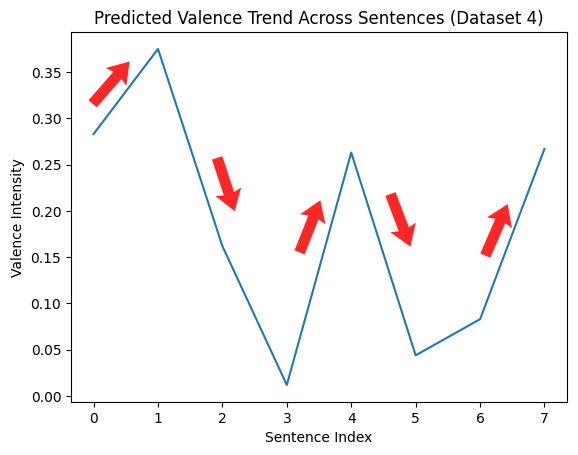}
    \caption{Predicted valence}
\end{subfigure}

\caption{
Comparison between ground-truth and predicted valence trajectories.
The predicted curves closely follow the temporal trend of the real values.
In Dataset 3, the negative dip and final increase are correctly reproduced.
In Dataset 4, the model captures the overall positive tendency and major
trend changes, although the predicted amplitudes are smoother than the
ground-truth values. These results indicate that the model can track
valence dynamics while slightly compressing extreme intensities.
}

\label{fig:valence_onecolumn}

\end{figure}

First, we use test samples from the CMU-MOSEI dataset to evaluate the affective perception module (see Figure~\ref{fig:valence_onecolumn}). Prediction accuracy for valence and arousal is assessed using mean squared error and mean absolute error metrics.

Second, the pedagogical reasoning module is evaluated based on instruction generation validity and inference latency. We specially measure whether the generated teaching-act representations meet predefined educational constraints and whether the reasoning latency satisfies real-time interaction requirements.

Third, we use quantitative motion metrics to evaluate the motion generation module, including mean displacement amplitude, velocity magnitude, root mean square jerk (RMS jerk), motion energy and spatial range computed from the BVH joint trajectories. These measurements assess gesture intensity, smoothness, and stability, and allow comparison across different teaching-act conditions to prove our model do distinguish them.

The results (see Table e~\ref{tab:motion_stats}) reveal that the explain condition leads to the highest amplitude, velocity, jerk and energy, showing more active and expressive gestures typically used during detailed explanations. The praise condition presents relatively low velocity and jerk but maintaining a large spatial range, corresponding to open but smooth congratulative gestures. What is more, the challenge, checkin, and neutral conditions exhibit moderate motion intensity with comparable amplitudes, reflecting controlled  conversational gestures commonly used during instructional interaction. The unclear condition shows increased variability compared to neutral motion, suggesting less stable and more irregular trajectories when the instructional intent is uncertain that aligned with how human deal with uncertain in real-world life. 

Overall, the quantitative results confirm that the proposed conditioning strategy generates physically feasible motion sequences that remain plausible for execution on a tuned humanoid robot, while preserving distinguishable pedagogical gestures styles across different instructional acts

\begin{figure}[t]
\centering

\begin{subfigure}[t]{0.49\linewidth}
    \centering
    \includegraphics[width=\linewidth]{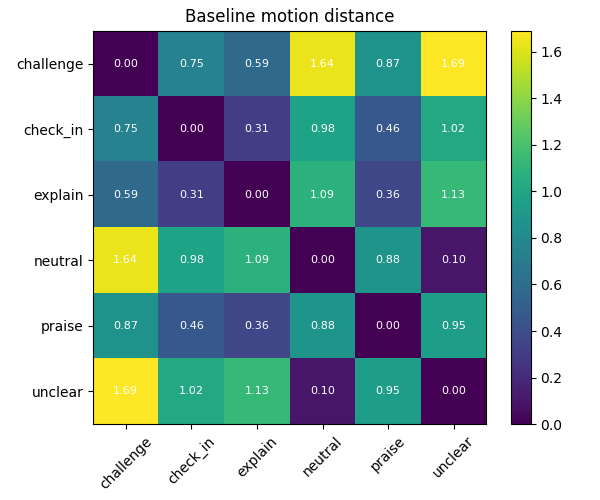}
    \caption{Baseline DSG+ without instructional conditioning}
    \label{fig:heatmap_baseline}
\end{subfigure}
\hfill
\begin{subfigure}[t]{0.49\linewidth}
    \centering
    \includegraphics[width=\linewidth]{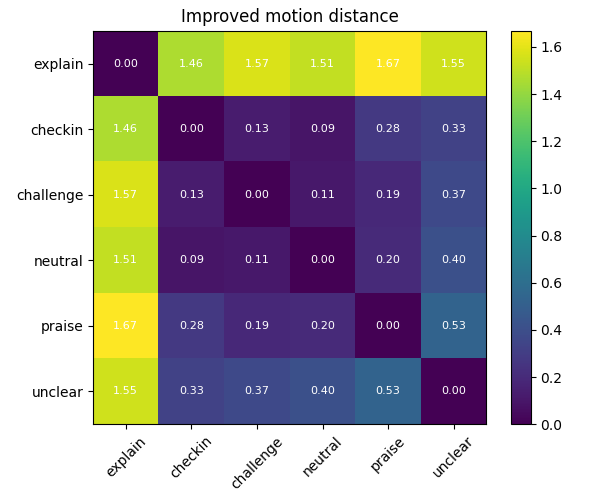}
    \caption{Proposed model with instructional conditioning}
    \label{fig:heatmap_improved}
\end{subfigure}

\caption{Pairwise distance heatmaps of normalized motion statistics across teaching-act categories. 
The baseline model shows weak separation between several acts, while the proposed conditioning produces a more structured distribution, with clear separation for highly expressive acts such as \textit{explain}.}
\label{fig:heatmap_compare}

\end{figure}

\subsection{Ablation Studies}
Ablation experiments are conducted to evaluate the effect of the proposed instructional conditioning in the RG-VLMD architecture.

First, we compare the proposed model with the original diffusion-based generator (DSG+), which does not use pedagogical intention signals. Motion statistics and pairwise distance heatmaps (Fig.~\ref{fig:heatmap_compare}) show that the baseline model produces weak separation across teaching-act categories, with several acts exhibiting very similar motion patterns, indicating limited control over gesture styles.

Then the proposed conditioning mechanism is enabled, where clip-level and frame-level teaching-act embeddings are added to the diffusion latent representation. Different from the original FiLM-based design that make the latent messy, the suggested additive conditioning with auxiliary act group supervision yields a more structured distribution of motion statistics. Particularly, highly expressive acts such as explain become clearly separated, while low-intensity conversational acts keep closer but still distinctive, revealing more consistent alignment between motion and instructional intent.  

Finally, removing frame-level scheduling results in more uniform gestures, while frame-wise conditioning facilitates smooth transitions between pedagogical behaviors within a single utterance. The final results confirm that reasoning-guided instructional conditioning improves controllability while maintaining physically plausible motion.


\section{Discussion}
The proposed Reasoning-Guided Vision-Language-Motion Diffusion framework demonstrates that integrating affective perception, pedagogical reasoning, and diffusion-based gesture generation enables humanoid robot to produce expressive and context-aware behaviors during Human-robot interaction (HRI) in educational scenarios. Different from conventional co-speech gesture motion generation systems that rely only on speech features, the proposed formulation explicitly involves learner affect and instructional intent into the gesture generation process. This design allows high-level pedagogical reasoning to guide the gesture synthesis rather than purely low-level acoustic cues.

The important observation from the experiments is that affective-driven educational conditioning significantly improves the consistency between generated gestures and the intended instructional strategy. When pedagogical control signals are removed, the diffusion model still outputs smooth motions but the gesture often lacks semantic relevance to the teaching context. This reveals that motion realism alone is not sufficient for educational interaction, and high-level reasoning signals are necessary to realize meaningful behavior in HRI scenarios.

Flexible control over gestural styles is provided by the hierarchical teaching-act representation through a combination of the global instructional vector and a frame-level schedule that enables smooth transitions between different pedagogical modes within a single interaction. It is particularly important to implement this temporal modulation in educational dialogue, where the robot may need to switch between explanation, encouragement, and guidance.  For the system, the modular design improves overall robustness and scalability, as the affect estimation, educational reasoning, and motion generation modules can be updated without retraining the entire system.
This design fits well into real-world deployment, where perception and language models may change over time, and the use of quantized and reduced diffusion steps further allows the motion generator to satisfy low-latency requirements while maintaining acceptable motion quality.

In the future, refinement work will focus on improving the adaptability of the system in real educational scenarios. Possible directions include incorporating reinforcement learning to optimize pedagogical decision-making, gathering diverse affective learning data sets, and extending the motion generation model to full-body control with stronger physical constraints. Additionally, user studies with students will be carried out at school to evaluate how empathetic gesture generation influences engagement, motivation, and learning outcomes.

\section{Conclusion}

This paper presents RG-VLMD, a reasoning-aware pedagogical diffusion framework for generating empathetic co-speech gestures for humanoid educational robots. The proposed system orchestrates multi-modal affective perception, language-based pedagogical reasoning, and teaching-act conditioned diffusion motion generation into a integrated architecture for real-time human-robot interaction. The robot is able to produce motion behaviors that are not only temporally aligned with speech but also consistent with the learner's affective state and the intended teaching strategy through the incorporation of valence/arousal estimation and instructional strategy representation.

The results of the experiment demonstrate that the semantic consistency and expressiveness of generated gestures compared to motion generation methods that rely only on speech conditioning are improved. The hierarchical teaching-act representation enables flexible control over gesture style, while the reasoning-guided conditioning allows the diffusion model to align motion according to pedagogical context. Additionally, the modular system design supports low-latency execution and facilitates deployment on a humanoid robot platform.

Although the current version relies on light-weight affect models and offline motion datasets, the results reveal that the reasoning-guided diffusion model provides a promising direction for building educational robot systems capable of adaptive, expressive, and context-sensitive interaction. Work in the future will investigate learning pedagogical policies from real classroom data, improving affect estimation in educational scenarios, and extending the motion model to full-body control with stronger physical constraints. 

In summary, this research demonstrates that combining affective computing, language reasoning, and diffusion-based gesture synthesis offers an effective system for developing next generation intelligent robot tutors.

\bibliographystyle{IEEEtran}
\bibliography{references}

\end{document}